# From Noise to Feature: Exploiting Intensity Distribution as a Novel Soft Biometric Trait for Finger Vein Recognition

Wenxiong Kang, *Member, IEEE*, Yuting Lu, Dejian Li, Wei Jia, *Member, IEEE*

*Abstract*—Most finger vein feature extraction algorithms achieve satisfactory performance due to their texture representation abilities, despite simultaneously ignoring the intensity distribution that is formed by the finger tissue, and in some cases, processing it as background noise. In this paper, we exploit this kind of 'noise' as a novel soft biometric trait for achieving better finger vein recognition performance. First, a detailed analysis of the finger vein imaging principle and the characteristics of the image are presented to show that the intensity distribution that is formed by the finger tissue in the background can be extracted as a soft biometric trait for recognition. Then, two finger vein background layer extraction algorithms and three soft biometric trait extraction algorithms are proposed for intensity distribution feature extraction. Finally, a hybrid matching strategy is proposed to solve the issue of dimension difference between the primary and soft biometric traits on the score level. A series of rigorous contrast experiments on three open-access databases demonstrates that our proposed method is feasible and effective for finger vein recognition.

*Index Terms*—finger vein recognition, soft biometric trait, vein imaging principal, intensity distribution

## I. INTRODUCTION

IN today's information and network society, personal authentication is becoming more important for security. There is no doubt that the technology of biometrics is one of the most effective solutions for this task [1][2]. In computer security, the technology of biometrics is defined as "automated methods of identifying or authenticating the identity of a living person based on a physiological or behavioral characteristic", which has shown significant advantages over traditional authentication mechanisms such as keys, passwords, personal identification numbers, and smart cards [1][2]. Usually, physiological traits, such as fingerprint, face, iris, palmprint, and hand geometry, are related to the shape of the body [1][2]. Behavioral traits, such as gait and voice, are related to the behavioral patterns of a person [1][2]. It is well known that fingerprint recognition, face recognition, and iris recognition are three representative biometrics technologies that have been studied in depth and widely used for different applications. In recent years, finger vein recognition has become an emerging biometric technology. In comparison with other biometrics, finger vein recognition offers significant advantages, such as high resistance to criminal tampering, living body identification, high speed of authentication, and compact device size. As a result, finger vein recognition is receiving increasing attention from academia and industry.

Traditionally, the fingerprint, face, iris, gait, voice, and finger vein are regarded as "hard" biometrics. To improve the performance of biometric systems, soft biometric traits were exploited [3][4][5]. Preliminary research on soft biometric trait was conducted by Jain *et al.* [3]. Recently, Dantcheva *et al.* [4] and Nixon *et al.* [5] surveyed the recent research progress on soft biometric traits. Generally, soft biometric traits can be defined as follows: soft biometric traits are physical, behavioral, or material accessories that are associated with an individual and that can be useful for recognizing an individual. These attributes are typically gleaned from primary biometric data, are classifiable in predefined human understandable categories, and can be extracted in an automated manner [4]. Compared to primary biometric traits, soft biometric traits may lack the distinctiveness and permanence to sufficiently differentiate any two individuals. However, they can provide some useful information about the individual [4][5]. Substantial research [6-8] has demonstrated that the performance of biometric systems can be improved by fusing primary and soft biometric traits.

According to [4] and [5], we know that most soft biometric traits are extracted from the face, body, iris, gait, and voice, and commonly used soft traits include gender and age. However, little work has been done to investigate the soft biometric traits in finger vein recognition. In the past research, only the finger geometry and the width of phalangeal joint were exploited as soft biometric traits for finger vein recognition [9-11]. As an emerging biometric modality, some of the potential characteristics of finger veins have not been revealed. In this paper, our objective is to investigate a novel soft biometric trait in finger vein recognition. After an extensive study and analysis, we find that the intensity distribution in the background of a finger vein image can be

Manuscript received on Jan 22, 2018. This work is partly supported by the National Science Foundation of China under Grants 61573151, 61673157; in part by the Guangdong Natural Science Foundation under Grant 2016A030313468; and in part by the Science and Technology Planning Project of Guangdong Province under Grant 2017A010101026 (Corresponding authors: Wenxiong Kang. email: auwxkang@scut.edu.cn)

Wenxiong Kang, Yuting Lu and Dejian Li are with the School of Automation Science and Engineering, South China University of Technology, Guangzhou, China.
Wei Jia is with the School of Computer and Information, Hefei University of Technology, Hefei, China (Email: china.jiawei@139.com).

treated as a novel soft biometric trait. Because different fingers have different structures, such as the position of the knuckle, the structure of the bone, the thickness of the finger, and the water content of the tissue, the intensity distributions of different finger veins are different after transmission imaging, which is often used for finger vein image acquisition. Although this intensity distribution was regarded as noise in past research [12-15], it contains some discriminant information. In this paper, we first analyze the imaging principle and explain why the intensity distribution in the background can be used as a novel soft biometric trait. Second, two background layer extraction algorithms and three soft biometric trait extraction algorithms are proposed for describing the intensity distribution. Finally, because there are significant differences in dimension between soft and primary biometric traits, existing matching strategies cannot achieve desirable performance; thus, we propose a hybrid matching strategy for fusing the primary and soft biometric traits for more robust finger vein recognition. This algorithm matches the primary biometric trait with SVM and the soft biometric trait with the Manhattan Distance. The results of a series of rigorous contrast experiments on three public databases indicate that our exploration of soft biometric trait extraction and matching is feasible and effective for a finger vein recognition system.

The main contributions of this work are as follows:

- We propose a novel soft biometric trait for improving the performance of finger vein recognition, *i.e.*, the intensity distribution in the background of the finger vein image. To the best of our knowledge, this is the first time that the intensity distribution has been investigated as one of the soft biometric traits of the finger vein.
- We propose an effective method for extracting the intensity distribution as a soft biometric trait. Because the finger vein image is composed of the foreground layer, which contains the texture information, and the background layer, which contains the intensity distribution information, the proposed method first uses a background layer extraction algorithm to separate the intensity distribution from the finger vein image. Then, the intensity distribution is described in three ways.
- The hybrid matching strategy is adopted to match the primary and soft biometric traits, which can further improve the matching performance. We conduct thorough experiments on three databases. Our method significantly outperforms previous state-of-the-art methods in terms of overall recognition performance.

The remainder of this paper is organized as follows: the related work on soft biometric traits is reviewed in Section II. In Section III, we propose a method for finger vein soft biometric trait extraction and introduce the details of the hybrid matching strategy. Experiments and our discussion are presented in Section IV. Our conclusions are presented in Section V.

## II. RELATED WORK

### A. Finger vein recognition algorithms

In the past two decades, many effective methods have been proposed for finger vein recognition, which can be divided into categories such as structure-based, local descriptor-based, subspace learning-based, correlation filter-based, orientation coding-based, and high level feature-based.

Structure-based methods usually extract structural features such as vein lines, the shape of the skeleton, curvature patterns, and minutiae, for recognition. Miura *et al.* [16] proposed a vein line-based method, in which line-tracking operations with randomly varied start points were repeatedly carried out. Later, Miura *et al.* [17] proposed a method for calculating the local maximum curvatures in cross-sectional profiles of a vein image, which can better extract vein lines. In [18], Huang *et al.* developed a wide line detector for feature extraction, which can obtain precise width information of the vein. Song *et al.* [19] designed a mean-curvature-based finger vein verification system, which viewed the vein image as a geometric shape and found the valley-like structures with negative mean curvatures. Liu *et al.* [20] presented a singular value decomposition (SVD)-based minutiae matching method for finger vein recognition. Considering that the vein networks consist of vein curve segments, Yang *et al.* [21] proposed a finger vein image matching strategy based on adaptive curve transformation. Yang *et al.* [22] proposed a structure-based method that includes an anatomical structure analysis-based vein extraction algorithm and an integration matching strategy. Recently, Qin *et al.* [23] proposed a deep representation-based feature extraction method that exploited a convolutional neural network to better segment vein patterns.

Local image descriptors are very popular methods in biometrics. For finger vein recognition, several local descriptors have been successfully exploited, such as local binary pattern (LBP) [24][25], local line binary pattern (LLBP) [26], efficient local binary pattern (ELBP) [27], local directional code (LDC) [28], and discriminative binary codes (DBC) [29].

For subspace learning-based methods, many dimensionality reduction techniques have been applied for finger vein recognition, such as principal component analysis [30], two-dimensional principal component analysis (2DPCA) [31], and linear discriminant analysis (LDA) [32].

Correlation filter-based methods, which are performed in the frequency domain, have shown promising performances for biometric recognition [33]. The attractive properties of correlation filters include shift invariance, noise robustness, graceful degradation, and distortion tolerance. Mahri *et al.* [34] and Yang *et al.* [35] proposed two effective algorithms that were based on band-limited phase-only correlation (BLPOC) for finger vein recognition.

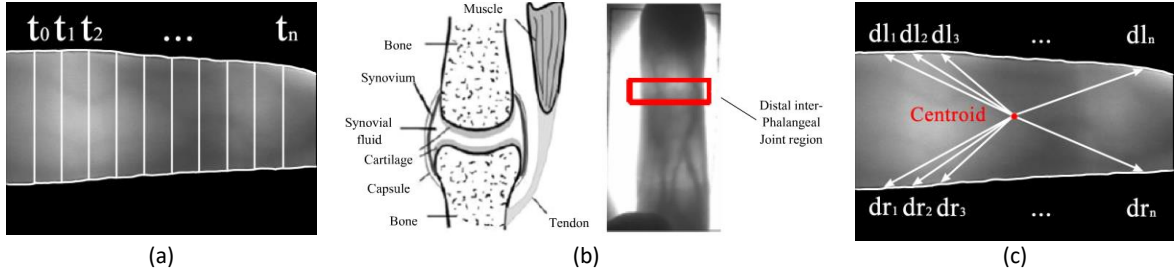

Fig. 1. Three reported soft biometric traits in finger vein images. (a) Finger shape as a soft trait of the finger vein; (b) the width of the phalangeal joint as a soft trait of the finger vein; (c) the distance between the finger edge to the image centroid as a soft trait of the finger vein.

Orientation coding-based methods have been widely used for palmprint recognition [33]. Usually, in these methods, the direction feature is quantized and represented in bits, and the Hamming distance is used to compare two palmprints. Yang *et al.* [36] proposed a comparative competitive coding method for finger vein recognition.

To further improve the recognition performance, many researchers utilized multi-modal strategies [37-41], in which the finger vein features and other biometrics were fused together to achieve higher precision.

### B. Soft biometric trait

As a novel research field on biometrics[5], the usage of soft biometric traits has been investigated in recent years. Jain et al. [3] preliminarily discussed and proposed a framework for the integration of soft biometric traits. Since then, soft biometric traits have been investigated in depth. Recently, Dantcheva et al. [4] and Nixon et al. [5] summarized the research details of soft biometrics and discussed the associated definition, benefits, applications, open research problems, and taxonomy. In [4], a taxonomy of soft biometrics was presented, which considered four groups of attributes: demographic, anthropometric, medical, and material and behavioral attributes. These four kinds of attributes, which are based on the modalities of the face, iris, body, gait, fingerprint, and hand, were comprehensively reviewed. The term demographics refers to attributes such as age, gender, ethnicity, and race, which are widely used in common population statistics [4]. Anthropometric attributes usually refer to geometric and shape features of the face, body, and skeleton [4]. Medical attributes are features that are related to body weight, body mass index, and skin color and quality [4]. Material and behavioral attributes are features that are related to eye lenses, glasses, hats, scarfs, clothes, color of the clothes, facial expressions, etc.

### C. Existing soft biometric trait of finger veins

Few works on the soft biometric traits of finger veins have been reported. Several soft biometric traits of finger veins that have been considered are shown in Fig. 1. In Fig. 1(a), Kang *et al*. [9] exploited finger shape as one soft trait to improve recognition performance. In Fig. 1(b), Yang *et al*. [10] employed the width of the phalangeal joint as a soft biometric trait to assist finger vein recognition and to increase robustness. In Fig. 1(c), Asaari *et al*. [11] proposed the distance between the finger edge to the image centroid as a soft biometric trait and achieved a satisfactory result.

### III. METHODOLOGY

As mentioned above, to take full advantage of the information in the finger vein image, we propose a new and effective soft trait of the finger vein based on the analysis of finger vein imaging theory. The flow chart of the finger vein recognition system framework combining a primary biometric trait and the proposed soft biometric trait is depicted in Fig. 2. First, we separate the input image into a foreground layer image and background layer image using our proposed image layer separation strategy. Next, the primary biometric trait is extracted from the foreground layer, and the soft biometric trait is generated from the background layer using different methods. Then, we proposed a hybrid matching strategy to match the primary and the soft biometric trait; that is, matching the primary biometric trait with SVM yields the main score, and matching the soft biometric trait with the Manhattan Distance yields the soft score. Finally, we normalize the two scores and fuse them based on the weighted

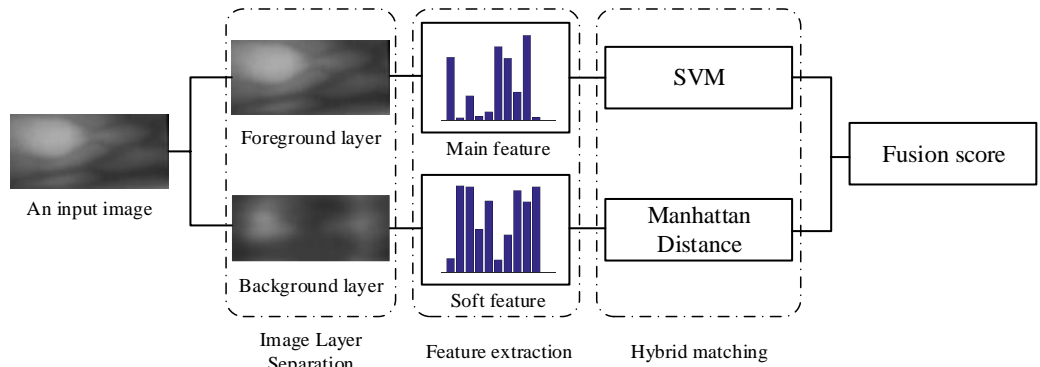

Fig. 2. Finger vein recognition system framework based on a primary biometric trait and a soft biometric trait.

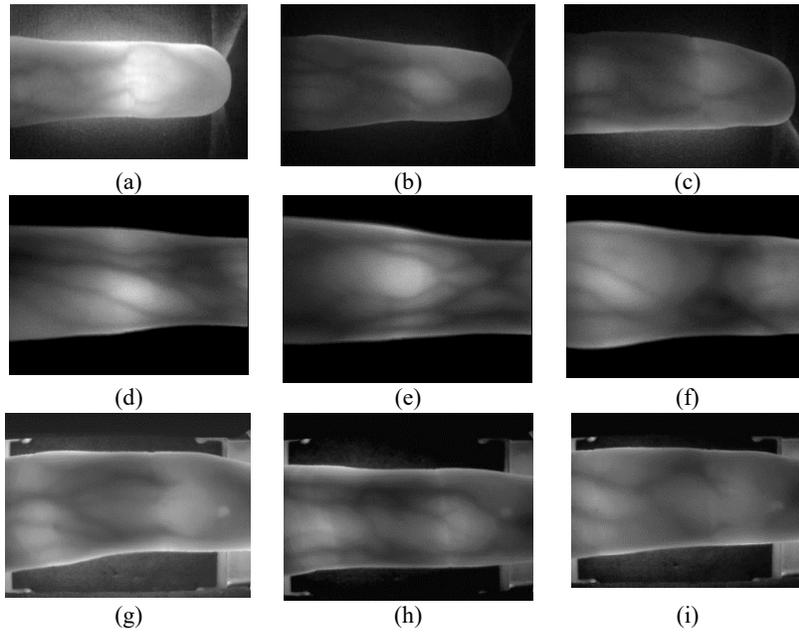

Fig. 3. Finger vein images from different databases. (a), (b), and (c) are from the FV-USM database; (d), (e), and (f) are from the MMCBNU database; (g), (h), and (i) are from the SDUMLA database.

sum rule to generate the final matching score.

*A. Finger vein imaging theory analysis*

Currently, there are two mainstream modes for finger vein image acquisition: transmission imaging mode and reflection imaging mode. However, despite the presence of self-adaptive illuminance control [42], it remains difficult to achieve a high-quality image in the reflection imaging mode because of its stringent requirements on the light source and the IR camera. Therefore, the transmission imaging mode is widely used in finger vein recognition devices. In this paper, we also focus on finger vein images that are obtained by the transmission imaging mode. In this mode, near-infrared (NIR) LEDs transmit rays through the finger. Some NIR rays are absorbed by the hemoglobin in the finger vein, while other NIR rays pass through other tissue and arrive at the camera, which projects a strip shadow to generate finger veins in the final image, as shown in Fig. 1(b). Owing to the distinction of genes, almost every individual has a unique finger vein texture. Thus, most of the finger vein feature extraction algorithms focus on only finger vein texture. However, based on a further observation, we found that the finger vein image includes not only the texture feature but also the feature of the intensity distribution, which is formed by the unique tissue constitution and distribution inside every finger. As shown in Fig. 1(b), the intensity of the knuckle region in the finger vein image is significantly brighter than that of other regions. Fig. 1(b) shows an image of the finger bone structure. Due to the low density of the synovial fluid, the NIR rays are more likely to penetrate the synovial fluid and generate a lighting block in the knuckle region of the finger vein image. This intensity distribution depends on the finger structure of the individual, such as the structure of the bone, the thickness of the finger, and the water content of the tissue. Based on the finger vein images of three open-access databases shown in Fig. 3 ((a), (b), and (c) are from the Finger Vein USM (FV-USM) Database [11], (d), (e), and (f) are from the MMCBNU_6000 Finger Vein Database [51], (g), (h), and (i) are from the SDUMLA-HMT Finger Vein Database [52]), not only the finger vein texture but also the intensity distribution in the background differs among individuals, which motivates us to explore the possibility of using the intensity distribution feature as a soft biometric trait in the paper. Although the proposed new soft biometric trait provides relatively weak discrimination, it is easily extracted, and the recognition performance can be effectively improved by fusing it with the primary biometric trait, which is demonstrated in Section IV.

*B. Background layer extraction*

As analyzed above, in addition to the texture feature, which is commonly included in finger vein images, the intensity distribution, which is formed by the unique tissue constitution and distribution inside every finger, can also be treated as a feature, which to the best of our knowledge has never been mentioned in existing literature. To extract the pure intensity distribution, the finger vein texture should be filtered out from the vein image, thereby leaving the neat background layer that is generated by other tissue. Since there have been no specific studies on it, we propose two algorithms for achieving this objective, i.e., image layer separation and Gaussian blur.

*1) Image layer separation*

Image layer separation (ILS) is a type of algorithm that separates an image into a smooth layer and a high-gradient layer, which is commonly used for intrinsic image decomposition and single-image reflection interference removal. Among all types of ILS algorithms, a relative smoothness-based method [43] is adopted as a comparison algorithm in this paper to solve the ILS issue for separating the finger vein image. This algorithm used the half-quadratic separation scheme and a normalization algorithm to separate a finger vein image into a foreground layer and a background layer. The foreground layer contains the texture information, and the background layer contains the intensity distribution information. Thus, the adopted ILS algorithm [43] is a valid method to separate the finger vein image. An example is given in Fig. 4, in which Fig. 4(a) shows three finger vein images selected from three different databases; Fig. 4(b) and (c) show the foreground layers and the background layers extracted by ILS, respectively.

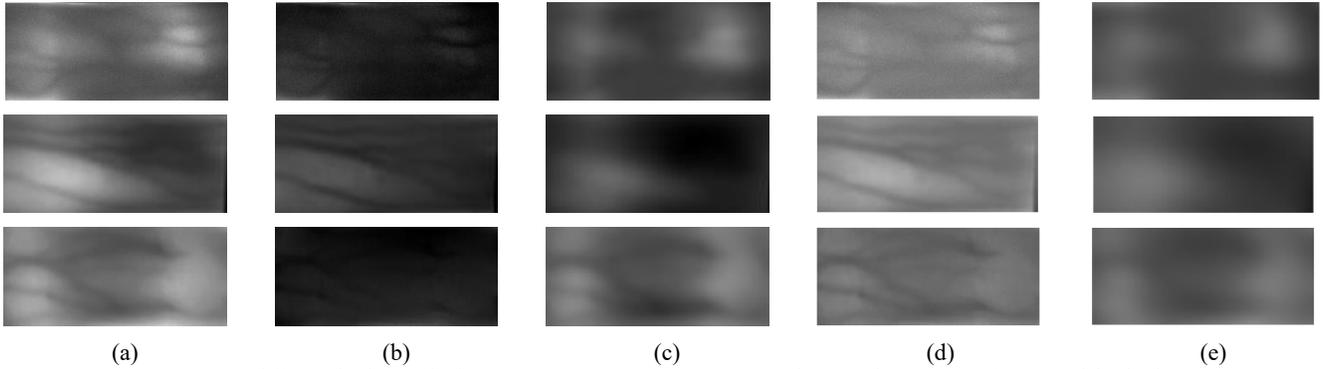

(a) (b) (c) (d) (e)

Fig. 4. Finger vein images from different databases: the first raw image of (a) is from FV-USM, the second is from MMCBNU, and the third is from SDUMLA; (b) foreground layers generated by ILS; (c) background layers generated by ILS; (d) foreground layers generated by GB; (e) background layers generated by GB.

*2) Gaussian blur*

Besides the ILS algorithm, Gaussian blur (GB) seems to be a more intuitive method to obtain the background layer. Because the finger vein texture can be viewed as the high-frequency components and the intensity distribution in the background can be regarded as the low-frequency components, GB can be used as a low pass filter to filter out the finger vein texture, thereby leaving only the background layer, which is illustrated in Fig. 5(d) and (e). As shown in Fig. 5(c) and (e), for the same finger vein image, the background layers that are obtained with IIS and GB are almost the same. However, GB is more efficient than ILS and has better noise suppression performance. To illustrate this, a series of rigorous contrast experiments on the effectiveness of GB and ILS for finger vein recognition will be presented in Section IV.

*C. Primary biometric trait extraction*

The primary biometric trait is the feature with the most powerful discrimination, which is key to ensuring the accuracy of the system. Many algorithms for finger vein primary biometric trait extraction have been introduced in Section II. Among them, LBP, Weber Local Descriptor (WLD), Histogram of Oriented Gradients (HOG), and Scale Invariant Feature Transform (SIFT) are the typical methods for achieving the described performance. In this paper, they are also used as the primary biometric trait extraction algorithms to evaluate the performance of our soft biometric traits.

LBP [44] is an efficient texture extraction algorithm with illumination and rotation invariance, which is usually applied to face recognition and vein recognition. The main idea of LBP is to measure the gray change between each pixel and its neighborhood and to code this change to generate the LBP code histogram, which can fully represent the finger vein feature but cannot handle the texture transition.

WLD [45] is also a popular texture extraction algorithm. Compared with LBP, WLD includes not only the gray information but also the gradient information. WLD calculates the gray-value difference between each pixel and its neighborhood, and measures the gradient information in the horizontal and vertical directions.

As an efficient texture extraction algorithm, HOG [46] was initially used for pedestrian detection. The major strategy of HOG is to divide the image into several cells, which are used to calculate the histogram of gradients. Then, the cells are divided into several blocks with overlapping regions to alleviate the impact of target rotation, which makes it possible for HOG to handle rotation around the finger in the finger vein recognition system.

SIFT [47] has been proven to be one of the most robust local feature descriptors in object recognition and matching. Difference of Gaussians was proposed for corner detection to improve the calculation efficiency, and the cardinal direction of the descriptor also makes it robust to object rotation.

*D. Soft biometric trait extraction*

According to the analysis in subsection III-A, the main difference among individuals' background layers is the intensity distribution. For this characteristic, we design three kinds of soft biometric traits that are extracted from the background layer, *i.e.*, the mean and variance, the array of mean and variance, and the histogram of spatial pyramid.

*1) Mean and variance*

The thickness of the finger and the density of the tissue vary among individuals. These physical characteristics are reflected in the brightness and the contrast of the finger vein image. Let $I(x,y)$ be the background layer image, and $w$ and $h$ be the width and the height of the image, respectively. Then, the grayscale mean $M$ is:

$$M = \frac{\sum_{i=1}^{wh} I_i(x,y)}{wh} \quad (1)$$

The variance $V$ is:

$$V = \frac{\sum_{i=1}^{wh} (I_i(x,y) - M)^2}{wh} \quad (2)$$

And the soft biometric trait of the mean and variance (M&V) is:

$$f_1 = [M, V] \quad (3)$$

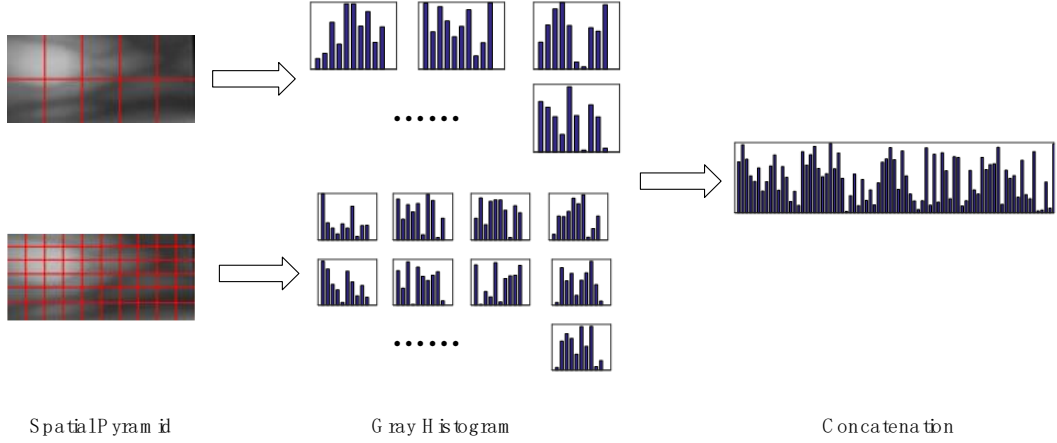

Fig. 5. Soft biometric trait extraction with the spatial pyramid.

*2) Array of mean and variance*

The width of the knuckle and the length of the finger vary among individuals. Therefore, the bright region and the dark region have different spatial properties among individuals. To describe these characteristics, the background layer image is divided into $a \times b$ blocks. The grayscale mean $m_i$ and grayscale variance $v_i$ ($1 \leq i \leq a \times b$) of each block are concatenated to obtain the $2 \times a \times b$-dimensional soft biometric trait, which is the array of mean and variance (AM&V):

$$f_2 = [m_1, v_1 \ldots m_i, v_i] \ (1 \leq i \leq a \times b) \quad (4)$$

*3) Histogram of spatial pyramid*

Soft biometric traits M&V and AM&V extract the features in only one scale. To construct the features in multiscale, a spatial pyramid [48] is used for sub-image partitioning, and the histogram is calculated in each sub-image. As shown in Fig. 5, the finger vein image is partitioned by two scale grids. Then, we calculate the grayscale histogram of each image block. The concatenation of these histograms in each block is called as Histogram of Spatial Pyramid (HSP).

*E. Hybrid matching strategy*

The primary and the soft biometric traits contain different amounts of information. If concatenating the primary and soft biometric traits directly from the feature level to construct the final biometric trait, the expected weight of the soft biometric trait and its discrimination may be reduced due to its lesser percentage in the final biometric trait. To address this problem, we propose a hybrid matching strategy at the score level. First, we adopt SVM to obtain the matching score for the primary biometric trait and use the Manhattan Distance to measure the similarity for the soft biometric trait. Then, we normalized and fused them with the weighted sum rule to generate the final matching score. The reason why different methods are adopted for the primary and the soft biometric traits is explained as follows: SVM requires only a few samples to obtain the optimal hyperplane and performs better in high-dimensional vector classification. Therefore, it is suitable for finger vein recognition with limited samples. Compared with the primary biometric trait, the soft biometric trait is a low-dimensional feature, so the Manhattan Distance is a more appropriate choice for soft biometric trait matching than SVM. Since the matching process with the Manhattan Distance is relatively simple, only the SVM matching process for primary biometric traits is introduced below.

SVM is commonly used for binary-class classification tasks, and multi-SVM can be used for multi-class classification tasks. However, the number of subjects is very large, while the number of samples of each subject is small in the biometric scenario. Therefore, regarding each subject as a class in multi-SVM makes it difficult to achieve the desired level of performance. Thus, we consider the finger vein recognition problem as a binary classification problem. Suppose $X_1$ and $X_2$ are the features for two samples and that their difference is $\Delta X = X_1 - X_2$. The intra-class difference $\Delta X_{positive}$ and the inter-class difference $\Delta X_{negtive}$ have different distributions and can be classified efficiently with SVM. In this paper, $\Delta X_{positive}$ is labeled as the positive sample and $\Delta X_{negtive}$ is labeled as the negative sample. The optimal hyperplane ($w$, b) can be obtained on the training dataset and then used to calculate the matching score:

$$\text{MatchScore} = w^T |\Delta X| + b \quad (5)$$

Here, $w$ and $b$ are the normal vector and the displacement, respectively. If $X_1$ and $X_2$ are intra-class samples, $\Delta X$ is labeled as 1 and $\text{MatchScore} \geq 1$. If $X_1$ and $X_2$ are inter-class samples, $\Delta X$ is labeled as -1 and $\text{MatchScore} \leq -1$. That is, the larger the value of $\text{MatchScore}$, the more similar the samples. In this paper, the LibSVM [49] toolbox is used to calculate MatchScore.

## IV. EXPERIMENTS

To evaluate the effectiveness of the proposed soft biometric trait and the hybrid matching strategy, four experiments are conducted in this section. The first is a comparison experiment of two different methods of background layer extraction: GB and IIS. The second experiment evaluates the effectiveness of the soft biometric trait on three open-access databases. The third experiment evaluates the performance of the hybrid matching strategy. Finally, the fourth experiment is a performance comparison between our algorithm and the state-of-the-art algorithms. The Equal Error Rate (EER) [50] is adopted in this section as the criterion for evaluating performance. EER is the error rate when the False Acceptance Rate (FAR) is equal to the False Rejection Rate (FRR). FAR is the probability that the system incorrectly authorizes a non-authorized finger vein. FRR is the probability that the system incorrectly rejects access to an authorized finger vein. Therefore, a lower EER of a finger vein recognition system indicates a better performance. All experiments were conducted on a PC with an Intel Core i5-4590 CPU with 64-bit Windows OS. The algorithms were programmed in Matlab 2015a.

*A. Databases*

Three open-access finger vein databases are utilized in this paper, including the Finger Vein USM (FV-USM) Database [11], the MMCBNU_6000 Finger Vein Database [51], and the SDUMLA-HMT Finger Vein Database [52].

*1) Finger Vein USM Database*

The images in the FV-USM database [11] were captured from 123 subjects, including 83 males and 40 females 20-52 years in age. Four finger veins were captured for each subject, including the left index, left middle, right index, and right middle fingers. Six samples were captured for each finger in each session. Every subject participated in two sessions, which were separated by an interval of more than two weeks. The whole database includes 5904 (123×4×6×2) samples. The resolution and depth of the images are 640×480 and 8 bit-gray-scale, respectively. To ensure a fair comparison, the ROI images that are provided by the database are utilized in this paper.

*2) MMCBNU_6000 Finger Vein Database*

The MMCBNU database [51] was captured from 100 subjects in 20 countries, including 83 males and 17 females 16-72 years in age. In this database, six finger veins were captured for each subject, including the left index, left middle, left ring, right index, right middle, and right ring fingers. Ten samples were captured for each finger, which resulted in a total of 6000 (100×6×10) samples. The resolution and depth of the images were 640×480 and 8 bit-gray-scale, respectively. The ROI images that are provided by the database are utilized in this paper.

*3) SDUMLA-HMT Finger Vein Database*

The SDUMLA-HMT database [52] was captured from 106 subjects, including 61 males and 45 females 17-31 years in age. In this database, six fingers were captured for each subject, including the left index, left middle, left ring, right index, right middle, and right ring fingers. Six samples were captured for each finger, which resulted in a total of 3816 (106×6×6) samples. The resolution and depth of the images were 320×240 and 8 bit-gray-scale, respectively. Since no ROI images are provided by the database, we first segment the ROI images in the database and utilize the segmented ROI images in our experiments.

*B. Comparison of background layer extraction algorithms*

This section presents a comparison experiment of two background layer extraction algorithms: ILS and GB. To present the results more clearly, only the FV-USM database is utilized in this experiment. The primary biometric trait is first extracted from the ROIs of the finger vein images that are provided by the database provider. Then, the soft biometric trait is extracted from the background layer images that are generated by the ILS and GB algorithms. For the primary biometric trait extraction algorithm, we proposed a gradient map-based rotation correction strategy to reduce the impact of rotation around the finger as follows: calculate the gradient map in the radial direction of one finger vein image as a reference map and move the other gradient maps from their upper edges to the lower edge of the reference map until the lower edges of these gradient maps coincide with the upper edge of the reference map. In this process, we make sure that the left and right edges of the images are aligned, and the Manhattan Distance of the overlapped gradient map is then calculated. The overlapping region that corresponds to the minimum distance is taken as the common region of the two images to be matched. Finally, the feature of the common region of finger vein images is extracted as the primary feature. Four primary biometric traits (LBP [44], WLD [45], HOG [46] and SIFT [47]), two background layer extraction algorithms (ILS and GB), and the three soft biometric traits that are mentioned in Section III-C are utilized in this comparison experiment. From the results in Table I, we can draw two conclusions. (1) Fusion of the soft and primary biometric traits can yield better performance than using the primary biometric trait alone, which demonstrates that the background layer of the finger vein image also contains useful identity information, which can lead to better discrimination among subjects. (2) The background layer that is extracted by GB can yield better performance than that of ILS. The reason is that, in addition to removing the finger vein texture, GB can also suppress noise and highlight the intensity distribution in the background layer, whereas ILS may lead to information loss and introduce noise when separating the texture and background layers. In terms of time consumption, the average elapsed time of GB is 4.58 ms, and that of IIS is 120.54 ms. Therefore, GB is utilized in the rest of the experiments in this paper.

TABLE I
COMPARISON OF BACKGROUND LAYER EXTRACTION ALGORITHMS (EER%)

|  | LBP | | WLD | | HOG | | SIFT | |
| --- | --- | --- | --- | --- | --- | --- | --- | --- |
|  | GB | ILS | GB | ILS | GB | ILS | GB | ILS |
| M&V | 0.257 | 0.271 | 0.799 | 0.840 | 0.447 | 0.433 | 0.257 | 0.352 |
| AM&V | 0.257 | 0.284 | 0.772 | 0.826 | 0.433 | 0.474 | 0.243 | 0.338 |
| HSP | 0.271 | 0.284 | 0.867 | 0.894 | 0.474 | 0.460 | 0.311 | 0.365 |
| Primary biometric trait only | 0.298 | | 0.921 | | 0.555 | | 0.433 | |

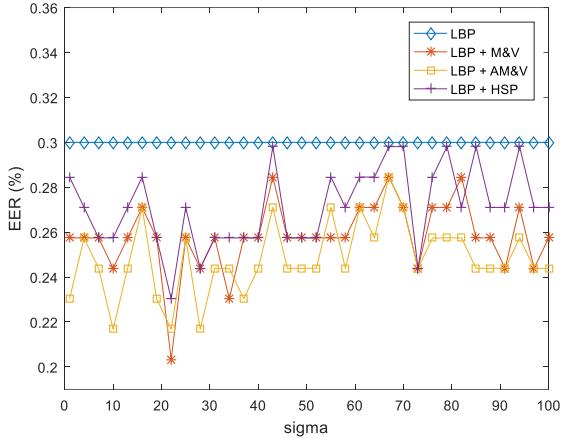

Fig. 6. Impact of sigma on soft biometric traits.

As mentioned above, GB can be used for removing the finger vein texture and highlighting the intensity distribution. The sigma parameter has a key impact on the generation of the background layer. If sigma is too small, the background layer still contains the texture feature, and the fusion of the primary biometric trait and the soft biometric trait may cause information redundancy; moreover, the noise cannot be suppressed efficiently. If sigma is too large, the image will also be blurred, which will remove a portion of the finger vein texture information and fade the intensity distribution. To evaluate the robustness of our finger vein soft biometric trait extraction algorithms, we adopt a series of sigma values and evaluate the recognition performance. To present the results more clearly, only LBP is used as the primary biometric trait. As shown in Fig. 7, the sigma value has certain impacts on the three soft biometric traits, especially on HSP, because it contains structural information that is vulnerable to over-blurring. Thus, the recognition result that is achieved by HSP is worse than the results achieved by other soft biometric traits. As shown in Fig. 6, when the value of sigma is approximately 22, the EER values that are obtained by the three soft biometric traits are the lowest among all sigma values. Therefore, sigma is set as 22 in the rest of the experiments. In addition, due to the performance of different block sizes in the array of mean and variance features and because the histogram of spatial pyramid features is different, the optimal block size in the array of mean and variance features and the optimal block size in the histogram of spatial pyramid features are set as 8*16 and 6*10, respectively, according to trial and error. Moreover, most of the results for the fused LBP and soft biometric traits are significantly better than those for LBP alone, which demonstrates that the soft biometric traits that are proposed in this paper are effective. Moreover, AM&V is more robust than the other two soft biometric traits because of the minimum fluctuation that occurs with changes in sigma.

### C. Recognition performance on different databases

This section presents the performances of the three soft biometric traits applied to three open-access databases: FV-USM, MMCBNU, and SDUMLA. FV-USM contains data from two sessions. For a more objective comparison, we adopt the same experimental protocol as that in the literature [11]: data from the first session are used to train the parameters, and data from the second session are used to evaluate the performance of our proposed method. The trained parameters are also directly applied to MMCBNU and SDUMLA. During the experiments, three soft biometric traits mentioned above are fused with four primary biometric traits, including LBP, WLD, HOG, and SIFT. To demonstrate the effectiveness of the soft biometric traits, four primary biometric traits are utilized separately in three databases. In Fig. 7, three of the curves correspond to the EERs of three soft biometric traits fused with the primary biometric trait and the other curve corresponds to only the primary biometric trait. The curves show that no matter which database we use for the experiment, fusion of the primary biometric trait and a soft biometric trait can achieve better performance relative to use of the primary biometric trait alone. Fig. 8 presents the EER curves of the four primary biometric traits that are fused with the three soft biometric traits on the three databases. All such cases demonstrate that taking full advantage of the background layer of the finger vein image can lead to improved performance by

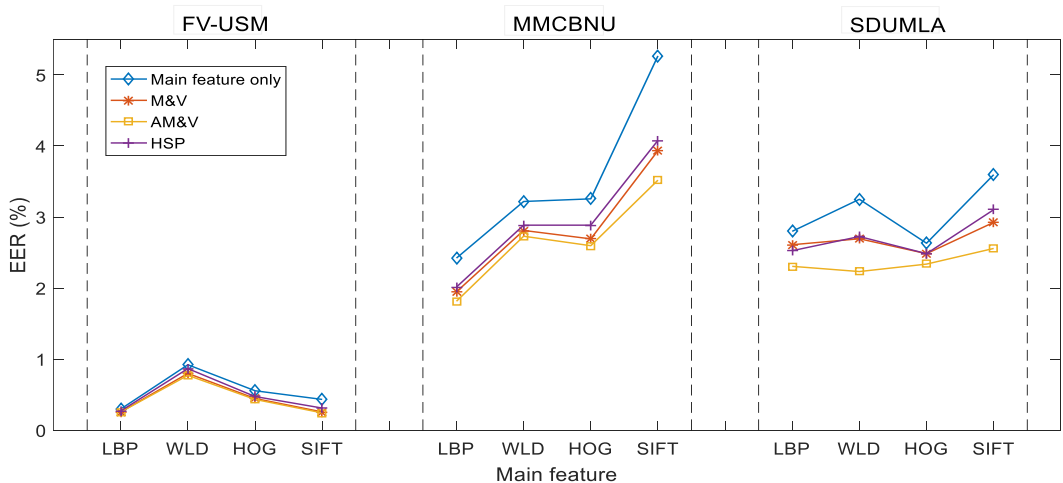

Fig. 7. Performances of soft biometric traits on three open-access finger vein databases.

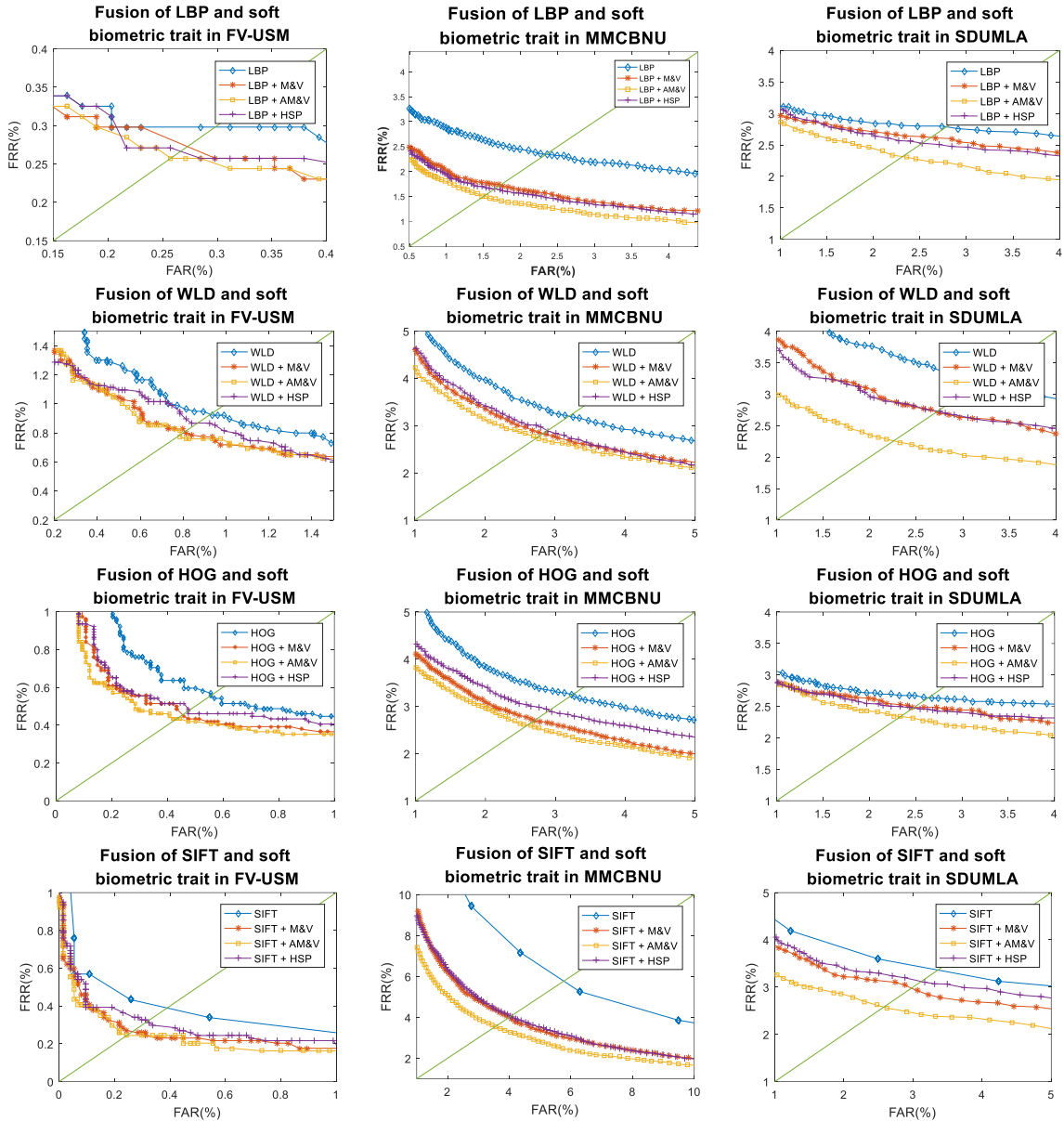
Fig. 8. EER curves of three databases.

combining a soft biometric trait with the primary biometric trait. Note that AM&V achieves better performance than the other soft biometric traits in most cases.

### D. Recognition performance of hybrid matching

This section examines the performance of the hybrid matching strategy. To present the results more clearly, we utilize two matching strategies for comparison, namely, the hybrid matching strategy (HM) and the single matching strategy (SM), which means that both the primary and soft biometric traits are matched with the Manhattan Distance. With two matching methods, LBP, WLD, HOG, and SIFT are utilized as primary biometric traits, and only the most stable soft biometric trait, namely, AM&V, is adopted. As shown in Table II, the EER of HM is significantly lower than that of SM. The primary biometric trait is represented by a high-dimensional vector, which is suitable for matching with SVM because the hyperplane of SVM can make excellent use of the information of high-dimensional vectors to obtain better classification results. Each soft biometric trait is represented by a low-dimensional vector. Therefore, it is more accurate and efficient to match the soft biometric traits with the Manhattan Distance than with SVM.

### E. Comparison with the state-of-the-art algorithms

Most papers on finger vein recognition have evaluated their algorithms on only one database and achieved desirable performance. However, evaluation on just one database is not objective and leads to inherent difficult in avoiding the problem of over-fitting. Therefore, evaluation on multiple databases is more reasonable. In the experiment, the session 1 data in FV-USM are used for parameter training, and the session 2 data in FV-USM, MMCBNU and SDUMLA are used for evaluation. A comparison of the EERs that are derived from recently published approaches on three open-access databases is given in Tables III-V, which indicates the superior performance of the proposed method. According to the results, our method achieves the best performance on three databases. This result is observed because most finger vein feature extraction algorithms focus on only finger vein texture and ignore the features in the background layer. To improve the performance of the finger vein recognition system, taking full advantage of the background layer is an effective strategy. Moreover, only the classical texture feature extraction algorithms are used as primary biometric traits in our paper, so it is possible that better performance can be achieved if more-advanced primary biometric traits are adopted.

TABLE II.
PERFORMANCE COMPARISON OF THE HYBRID MATCHING STRATEGY AND THE MANHATTAN DISTANCE (EER %)

|            | FV-USM |       | MMCBNU |       | SDUMLA |       |
|------------|--------|-------|--------|-------|--------|-------|
|            | SM     | HM    | SM     | HM    | SM     | HM    |
| LBP+AM&V   | 0.257  | 0.216 | 1.851  | 0.669 | 2.306  | 1.656 |
| WLD+ AM&V  | 0.772  | 0.379 | 2.729  | 0.94  | 2.232  | 0.755 |
| HOG+ AM&V  | 0.433  | 0.108 | 2.596  | 0.449 | 2.337  | 0.922 |
| SIFT+AM&V  | 0.243  | 0.237 | 3.518  | 2.763 | 2.557  | 0.712 |

TABLE III.
COMPARISON WITH THE STATE-OF-THE-ART ALGORITHMS ON FV-USM

| Reference | Year | Methodology | EER% |
|---|---|---|---|
| Qiu et al. [31] | 2016 | Dual-sliding-window localization and pseudo-elliptical transformer | 2.32 |
| Asaari et al. [11] | 2014 | Band-limited phase-only correlation | 2.34 |
| Wang et al. [53] | 2014 | Gabor wavelet features | 4.75 |
| Qin et al. [23] | 2017 | Deep representation-based feature extraction | 1.69 |
| Proposed method | 2018 | Combining primary and soft biometric traits | **0.216** |

TABLE IV.
COMPARISON WITH THE STATE-OF-THE-ART ALGORITHMS ON MMCBNU

| Reference | Year | Methodology | EER% |
|---|---|---|---|
| Lu et al. [54] | 2014 | Histogram of salient edge orientation map | 0.9 |
| Meng et al. [28] | 2012 | Local directional code | 1.03 |
| Xie et al. [55] | 2015 | Guided-Filter-Based Singe-Scale Retinex | 1.5 |
| Proposed method | 2018 | Combining primary and soft biometric traits | **0.827** |

TABLE V.
COMPARISON WITH THE STATE-OF-THE-ART ALGORITHMS ON SDUMLA

| Reference | Year | Methodology | EER% |
|---|---|---|---|
| Qiu et al. [31] | 2016 | Dual-sliding-window localization and pseudo-elliptical transformer | 1.59 |
| Wang et al. [53] | 2014 | Gabor wavelet features | 2.36 |
| Vega et al. [56] | 2014 | Morphologic operation of skeletonization | 27.56 |
| Liu et al. [20] | 2014 | Singular value decomposition-based minutiae matching | 2.46 |
| Xi et al. [29] | 2017 | Discriminative binary codes | 0.88 |
| Proposed method | 2018 | Combining primary and soft biometric traits | **0.712** |

V. CONCLUSIONS

In the past, most research of finger vein recognition only focuses on the texture feature of finger veins but gives little attention to the intensity distribution in the background, even regarding the intensity distribution as the noise. This paper analyzes the theory of finger vein imaging and the features in the image and proposes a soft biometric trait extraction algorithm. First, the background layer without finger vein texture is extracted with ILS and GB. Then, the intensity distribution in the background layer is described by three soft biometric traits. Finally, a hybrid matching strategy for improving the matching accuracy of the primary biometric trait and the soft biometric trait is proposed. The experimental results showed that GB is less time-consuming than ILS but achieves equivalent performance. Moreover, the results, when applied to three open-access databases, showed that fusion of the primary biometric trait and the soft biometric trait can yield a lower EER than when using only the primary biometric trait, which demonstrates the efficiency and universality of the proposed soft biometric trait. The performance of the soft biometric trait remains stable over a series of sigma changes, which demonstrates that the soft biometric trait in the background layer is robust. In this work, we presented a preliminary study of the soft biometric trait based on the intensity distribution. In our future work, we will continue to focus on this issue to better understand finger vein patterns and will propose more effective features to further improve the recognition performance.